\documentclass[leqno,fleqn,12pt]{article}

\usepackage{latexsym}
\usepackage{avm}
\usepackage{graphicx}
\usepackage[T1]{fontenc}
\usepackage{natbib}
\usepackage{array} 
\usepackage{todonotes}

\avmfont{\scriptsize\sc} \avmsortfont{\tiny\it}

\newcounter{corr}
\stepcounter{corr}

\title{Composing or Not Composing?\\Towards  \emph{Distributional Construction Grammar}}

\author{Philippe Blache$^{1, 5}$ \and Giulia Rambelli$^2$ \and Emmanuele Chersoni $^3$ \and Alessandro Lenci$^4$}

\date{\small $^1$Laboratoire Parole et Langage, CNRS, France \\ $^2$ Department of Languages, Literatures and Modern Cultures, University of Bologna, Italy \\ $^3$ Department of Chinese and Bilingual Studies, The Hong Kong Polytechnic University, Hong Kong  \\ $^4$ Department of Philology, Literature, and Linguistics, University of Pisa  \\ $^5$ ILCB -  Institute of Language, Communication and the Brain, France}

\usepackage{setspace}
\begin{document}

\maketitle
\begin{abstract}
The mechanisms underlying language comprehension remain an open question. Classically, building the meaning of a linguistic utterance is said to be incremental, step-by-step, based on a compositional process. However, several works have long demonstrated that non-compositional phenomena are also at work. It is therefore necessary to implement a unified approach that integrates both approaches.

In this paper, we propose a framework based on \textit{Construction Grammar} that accounts for these different access mechanisms. First, we provide a formal description of this framework by extending the feature structure representation proposed in \textit{Sign-Based Construction Grammar}. We present a general representation of the meaning based on the interaction of three semantic components: constructions, frames and events. Secondly, we describe how to implement a processing mechanism for building the meaning based on the notion of activation evaluated in terms of similarity and unification.


\end{abstract}


\newpage
\section{Introduction}


This paper addresses a long-standing and still unresolved question: What are the cognitive and linguistic principles that underlie natural language comprehension? While most existing research has focused on written texts and formal language use, we still lack a clear understanding of how meaning is accessed in more natural settings such as everyday conversation where speakers often produce incomplete, noisy, or non-canonical utterances.

The classical view is grounded in the principle of compositionality (see \cite{szabo2004compositionality} for an overview) which holds that the meaning of a sentence or more generally a message, whether written or spoken, is a function of the meanings of its constituent parts. This process is typically understood to be incremental: individual word meanings are combined into increasingly complex structures in a stepwise, hierarchical fashion until a full interpretation is achieved.

However, research in theoretical, descriptive, and experimental linguistics has shown that meaning can often be derived more holistically from global patterns or conventionalized structures without relying solely on compositional mechanisms. The large number of facilitation effects observed in processing idiomatic expressions and other multiword expressions \citep{contreras2022models}, as well as compositional frequent expressions \citep{rambelli2023frequent}, goes against a strong version of compositionality. Indeed, we may still recognize the parts composing the meaning of well-known expressions (i.e., their compositionality), without this entailing that we ``build them from scratch'' each time we use them.
At the same time, several studies demonstrate that online processing may occur in a \textit{good-enough} manner, relying on expectations, and potentially resulting in shallow interpretation or misinterpretation, such as for garden-path sentences \citep{christianson2001thematic,ferreira2003misinterpretation} or semantic attraction \citep{kim2005independence,Brouwer17,cong2023language}. Moreover, efficient comprehension usually adopts contextual constraints to anticipate or predict upcoming input, leading to facilitated processing once the expected component is encountered \citep{levy2008,huettig2015,Ferreira2016-ve,pickering2018predicting,pickering2013integrated,goldberg2022good,Blache24,britton2024influence}.


Taken together, this body of evidence suggests that two distinct but concurrent mechanisms are involved in language comprehension. The first is a \textbf{compositional} route, in which meaning is constructed incrementally, word by word, as each new lexical item is integrated into an evolving syntactic and semantic structure. The second is a \textbf{non-compositional} route, which relies on the recognition of familiar patterns or constructions, enabling more direct access to meaning without full syntactic parsing. A central challenge, then, is to develop a unified formalism that can account for the interaction between these two modes of processing.



Additionally, when trying to explain the understanding process, a second important question is the influence of the context. More precisely, several approaches in sociolinguistics, corpus linguistics and cognitive linguistics \citep{Langacker00,Tomasello05,Bybee10} have underlined the fact that meaning emerges from language use. These so-called \textit{usage-based} theories argue that every linguistic structure is shaped by repeated exposure and usage patterns.  
In particular, at the lexical level, the \textbf{distribution of the words} constitutes one of the most important sources of information for accessing their meaning: usage specifies the semantic relations between words.

To summarize, explaining language comprehension needs to account for the fact that 1) meaning can be accessed following two different routes: incrementally, in a compositional way, but also  globally, based on a direct access and 2) meaning depends on use and context.

While traditional theories do not account for these phenomena, these observations are at the basis of \emph{Construction Grammar} \citep{Fillmore88,Goldberg03,HoffmanTrousdale2013,Hilpert2019,Hoffmann2022}. 
In this theoretical paradigm, linguistic objects are organized in terms of form/meaning pairs called \textit{constructions} (hereafter \textit{Cx}; \citealp{Goldberg03,Goldberg06}). While there is no converging definition and operationalization of Cxs \citep{ungerer2023constructionist}, what is interesting is that constructions can be elementary objects (morpheme, words) as well as complex ones \citep{Goldberg03,Goldberg06,Sag12}: multiword expressions, idioms, syntactic constructions (e.g., ditransitive, passive, covariational conditional etc.). Moreover, constructions differ in their degree of schematicity, ranging from fully lexicalized and fixed expressions (e.g., \emph{kick the bucket}) to highly schematic and fully productive constructions (e.g., the \textit{ditransitive construction}). In all cases, the form-meaning pairing is at the core of the definition of a \emph{Cx}: recognizing the form makes it possible to access its meaning directly. For example, in the case of idioms, as soon as the construction is identified (typically after 2 or 3 words), the complete meaning is accessed directly, without any compositional mechanism. 
However, most constructionist approaches have primarily focused on formalizing representations rather than modeling the dynamic processes of language comprehension. To bridge this gap, a more comprehensive framework is needed that 1) operationalizes how Cxs are integrated during processing by means of compositional and noncompositional mechanisms, 
and  2) includes usage-based semantic information into structured linguistic descriptions.

This paper addresses both issues with the final aim to build a constructionist model of language processing. 
Considering the first point, it is necessary to identify in which situation one route is used instead of the other. Our hypothesis is that, by default, a pattern-based mechanism is applied, making it possible to recognize constructions and explaining different shallow processing effects \citep{Ferreira07}. In this case, constructions are recognized on the basis of different cues  (morphological, lexical, distributional, etc.). This hypothesis relies on the notion of \textit{activation} \citep{Anderson04,Lewis05,Blache24}, considering that a construction can be recognized above a certain activation threshold. 
Considering the second issue, most examples of constructional meaning are hand-coded, leaving the semantics of constructions unclear \citep{feldman2020advances,ungerer2023constructionist}. To overcome this limit, we introduce a novel semantic representation of Construction Grammar, named \textit{Distributional Construction Grammar} (DCxG). The idea is to incorporate distributional vectors into the construction representation, thereby accommodating the more usage-based aspects of meaning.
To summarize, the goal of this paper is twofold:

\begin{itemize}
\itemsep=-1pt
\item Specifying the nature and the representation of constructions, integrating usage-based models of meaning  (i.e., the \textit{Distributional Semantics} framework);
\item Defining an activation function and the different mechanisms making it possible to identify when constructions are recognized.
\end{itemize}

\noindent{}The perspective of this paper is mainly theoretical, with the aim of laying down the foundations of a general model of meaning construction to be used to guide experimental work for cognitive modeling of language processing.

\section{Situation}

\subsection{Formal and Distributional Semantics}
Formal Semantics addresses the questions of knowledge representation and mechanisms for building meaning. In many cases, semantic information is represented in terms of feature-value structures, and logic is used both as a descriptive language and as a calculus for meaning construction, thanks to different types of inference relations \citep{partee2011formal}. In these approaches, meaning is built starting from atomic entities (typically the meanings of words) that are incrementally combined into larger structures. Such a mechanism constitutes the basis of compositionality, for which the meaning of an expression is a function of the meaning of its constituents. This approach relies on an explicit representation of both types of information: the one associated with the constituents (typically lexical semantics) as well as the inference mechanisms (typically the inference rules). 

On the other hand, Distributional Semantics \citep{Lenci08,Lenci:Sahlgren:2023} aims at representing usage-based information directly. In this approach, the meaning of a word is derived from the contexts in which it occurs in different corpora and represented by means of a distributional vector. The notion of context, as well as the type of information taken into consideration, can vary: basic distributional models use the distribution of lexical forms, while more complex approaches propose to take into consideration sequences of forms, as well as richer information such as syntax \citep{Lenci08, Baroni10,clark2015vector}. In such approaches, the distribution of the observed linguistic items provides the basis for the notion of meaning.


Bringing together formal and usage-based approaches constitutes a major challenge for linguistics in general and for semantics in particular. Moreover, this question also crosses cognitive aspects: What type of meaning is stored in the brain? How is it accessed? How does comprehension work?

These questions are addressed differently depending on the theoretical and the operational framework. In formal approaches, it is assumed that meaning emerges from the interaction of different linguistic domains. In the Montagovian framework \citep{Montague73}, a homomorphism mapping from syntax to semantics is the basis of the construction of linguistic meaning. On the opposite side, for distributional approaches, meaning mainly comes from the context: without requiring any specific combinatorial operation, pieces of meaning, at the basis of understanding, can be accessed through the usage itself. Clearly, the distinction between these approaches reflects the Chomskyan opposition between \emph{i-language} and \emph{e-language} \citep{chomsky1980rules}, which would, in theory, result in irreconcilable views, explaining partly their mutual ignorance until recently. Moreover, this opposition has been reinforced by the fact that this distinction also recovers an opposition in the processing techniques: Formal Semantics is mainly implemented through symbolic and rule-based approaches, whereas distributional knowledge is numerical and probabilistic. On the one hand, the first approach is \emph{expert-based} and requires the explicit encoding of linguistic knowledge (grammars, inference rules, etc.). On the other hand, the processing mechanisms behind distributional models are purely \textit{data-driven}: 
basically, no prior linguistic knowledge is required, and all information emerges from data. Without reaching what can be called \textit{understanding}, this second track has led to several applications, for example, in information retrieval. 

It is important to underline the respective advantages of the two approaches. The former aims to explain how meaning is built by providing a formal framework, relying on a general theory of meaning, and making it possible to represent and process information. The latter provides tools dealing with co-occurrences between linguistic forms, starting from the concrete usage of words and expressions in natural linguistic contexts. It does not necessarily aim at constituting a theory of meaning,\footnote{However, see \cite{Lenci08} for a distinction between a \textit{weak} and a \textit{strong} interpretation of distributional models. According to the latter, the distributions of linguistic items may have a causal role in shaping (at least part of) the meaning representations, rather than being simply correlated with meaning.} but rather to capture elements of meaning at a large scale. Each perspective has its own goals, drawbacks and advantages, and we propose to reconcile them. Several works in the Natural Language Processing community have recently gone in this direction typically by addressing the question of word combination in vector space \citep{Mitchell10a, Grefenstette13,Lewis13,Baroni13,Boleda16,emerson2016functional,emerson2018functional,lo2023functional} or integrating both sources of information at the propositional level \citep{Beltagy16}.

In their proposal, \cite{Asher16} define how to integrate distributional semantics into \textit{Type Composition Logic} (TCL). This approach is based on the  representation of TCL's notion of internal content with vectors, integrating a distributional representation of word meaning into a model-theoretic approach. Internal content plays the role of a constraint controlling composition. This work proposes a way to integrate the two types of information, letting aside the question of a joint representation. 

More recently, \cite{Venhuizen22} have proposed the framework of \textit{Distributional Formal Semantics}, addressing exactly this goal of bringing together into a unique and homogeneous representation distributed meaning representations and proposition-level meanings. More precisely, DFS aims to gather propositional meaning in terms of models with lexical meaning in terms of linguistic context; the meaning remains organized around a proposition-central and compositional perspective.

\subsection{Accessing to Meaning: (non-)Compositionality}

The question of compositionality crosses, at least partly, the distinction between formal and distributional semantics. Formal semantics and its associated calculi consider that the meaning is formed by elementary components that are assembled incrementally, along the sentence, text, and discourse. This mechanism first identifies the basic elements and then builds abstract representations aiming at capturing the meaning of complex structures, following the classical Fregean principle. For example, first-order logic can be applied basically by identifying the arguments and the predicates and gathering them into formulae, thanks to a mapping function from syntax to semantics \citep{Montague73}. Quantifiers and modalities complete the structure, following specific mechanisms based on more complex calculi. This general mechanism is more or less the same across different theories and relies on two assumptions: 1) Meaning can be decomposed into atomic elements, and 2) there exists a linear and incremental mechanism assembling these elements into abstract structures. These assumptions are at the basis of many formal semantic frameworks, in particular those focusing on the syntax/semantics/discourse interface, such as the \textit{Discourse Representation Theory} \citep{Kamp93} or \textit{Combinatory Categorial Grammars} \citep{Steedman00}. From a computational perspective as well, natural language understanding approaches have also relied for decades on the Montagovian view of compositionality. Extensive work has been conducted in this direction within the logic programming paradigm \citep{Colmerauer82,Shieber87}, and, more recently, within \textit{Categorial Grammars} \citep{Bos04, Moot12}. 
Another framework exploring semantic representation with minimal structures has been proposed within the HPSG paradigm: \emph{Minimal Recursion Semantics} \citep{Copestake01} introduces important notions such as \textit{underspecification} and direct interfaces between semantics and other domains, on top of syntax.

These different approaches, both theoretically and computationally, not only offer precise descriptions of many phenomena, but also provide a precise framework for implementations in the domain of natural language understanding. 

However, compositionality has been questioned more recently, based on evidence from specific linguistic phenomena in which meaning is not the result of the composition of sub-constituents. This is typically the case when processing entire language patterns that can be recognized. The nature of such patterns and the mechanisms underlying their formation are central questions in linguistic theory. Usage-based approaches, particularly those proposed by \cite{Bybee06,Bybee13}, address these questions through the concept of chunks that are multi-word sequences whose formation is grounded in usage and frequency. Chunks are frequently used expressions that are stored and processed as whole units in the mind. Crucially, rather than being parsed word by word, they are accessed holistically, much like single lexical items. This holistic processing is often accompanied by a loss of analyzability, meaning that the internal structure of the chunk becomes less transparent.

Frequency plays a key role in the emergence of chunks: the higher the co-occurrence frequency of a sequence of words, the stronger the cognitive association between them. Through repetition, such sequences become automatized, making their processing and integration more efficient compared to compositional, word-by-word processing. Over time, this leads to a reduction in grammatical transparency within the chunk.

In Bybee's framework, grammar itself emerges from chunking. Grammatical patterns, such as verb conjugations or auxiliary constructions, originate as frequently used sequences that become schematized into productive constructions. The notion of construction, which underlies usage-based grammar, is intimately tied to chunking: just as fixed sequences can become grammaticalized, more abstract patterns can emerge through similar processes, resulting in constructions that reflect entrenched usage.

This is typically the case with idiomatic constructions \citep{Sag02, Baldwin10}, in which their figurative meaning cannot be accessed through a compositional process, and multi-word expressions \citep{Goldberg16,contreras2022models}. However, there are several linguistic phenomena which bear a meaning that cannot be decomposed  \citep{Culicover05, Goldberg16}. In the Construction Grammar literature, a classic example is the comparative correlative construction (\citealp{HoffmannBrunnerHorsch2020}; also ``Covariational Conditional'', cf. \citealp{culicover1999view}) in the form of ``\emph{the Xer, the Yer}'' (for example ``\emph{The larger the audience, the easier the show.}'') can only be interpreted because meaning is directly associated to the construction, \emph{as a whole}. There is no decomposition into sub-constituents that can be linearly assembled: The general interpretation comes only from the fact that two comparatives are juxtaposed. This suggests that, in many cases, meaning is accessed directly rather than being constructed step by step. 
While these linguistic phenomena are problematic for traditional theories of language in general and formal semantics in particular, these constitute the basis of the notion of construction in \emph{Construction Grammar} \citep{Hoffmann2022}. In many cases, interpretation relies on the interaction of multiple sources of information, including non-linguistic cues such as intonation, gazes, gestures, and body posture. Constructions are thus considered as form/meaning pairs, in which the global interpretation can only be accessed directly, even though some subparts may be analyzed compositionally. In other words, interpreting a construction consists mainly in recognizing its form, the meaning being directly accessible.

Therefore, we observe two engines that drive interpretation. On the one side, the compositional construction of meaning relies on the identification of the different semantic constituents and their assembling. This clearly relies on a step-by-step mechanism in which each stage consists in integrating incrementally partial information into a larger structure, this mechanism being controlled by syntactic information. The natural (and classical) implementation of this approach is \textit{grammar-driven}, exploiting an explicit representation of knowledge by means of rules, principles, constraints, etc. On the other side, non-compositional mechanisms,  typically involved in the interpretation of constructions, rely on a global recognition: the form of construction represents a pattern, which is specific enough to be identified in itself, without needing the analysis of its syntactic organization\footnote{It is worth mentioning that compositional operations and a construction-based formalism to syntax are not inherently contradictory, and some constructionist approaches have proposed how to integrate them into a unique formalism \citep{kay2012constructional}.}. Concretely, in such conditions, there is no particular mechanism for incrementally building the meaning, only a recognition of the general pattern that mainly relies on the accumulation of different cues at any level: words, gestures, context, etc. In such case, the problem consists first in identifying what a cue is and how cues interaction can trigger the recognition of a pattern. 

Therefore, in sentence comprehension, the access to meaning can take this double route, sometimes exploiting compositionality, sometimes with direct access. Such hypothesis relies on the idea that both mechanisms are at work in natural language understanding: \emph{A compositional process, unless direct access to the meaning becomes available thanks to a form/meaning coupling}. 
This distinction between compositional and non-compositional mechanisms crosses the debate between classical dual-system (relying on a clear distinction between lexicon and grammar) and single-system models of language processing (with no distinction between stored and computed forms) \citep{Ullman99}. For the latter, this information is learned and comes from linguistic experience, as proposed in usage-based approaches \citep{Bybee10}. 

Behavioral and neurocognitive evidence supports the existence of both processes (cf. \citealp{Rambelli2024} for a review of the experimental literature about non-compositional access). On the one hand, many fundamental works have shown the existence of an incremental integration, underlining the role of syntax in the process leading to comprehension, and then the compositionality of the mechanism \citep{Hagoort03,Friederici11,Berwick13, Ding16}. On the other hand, non-compositional mechanisms based on a direct form/meaning pairing have also been observed in language processing, relying on the memorization of entire structures or networks of connections \citep{Ullman01,  Kounios01,Holsinger13,Pulvermuller13}. 
From the perspective of cognitive neuroscience, \cite{Baggio2021} argues that the human brain can process language both compositionally and non-compositionally. Specifically, he claims the existence of two parallel streams, a semantic system and a grammar system: a semantic interpretation may be generated by a syntax-driven stream and by an ``asyntactic'' processing stream, jointly or independently. This theory has been supported by behavioral and neurological studies and has been formalized inside the theoretical framework of Parallel Architecture \citep{Jackendoff:1997}, which share similar assumptions about language with Construction Grammar  \citep{jackendoff2013constructions}.


To summarize, access to meaning can be both compositional and non-compositional, depending on the constructions, the sources of information, and the communicative situation. Moreover, in many configurations, the two mechanisms are involved together. Typically, many constructions are flexible in the sense that they contain variable components. For example, constructions such as ``\emph{the Xer, the Yer}'' or ``\emph{what's X doing Y}'' \citep{Kay99} have an intrinsic non-compositional meaning. But their complete understanding requires the interpretation of their variable parts $X$ and $Y$, that can be done compositionally, depending on their form. In other words, the meaning of a construction and, more generally, a complete sentence can be accessed by means of both mechanisms at the same time. It is then necessary to propose a hybrid theoretical framework integrating both aspects.

\subsection{Knowledge of the World: Frames and Events}
Besides the mechanisms for accessing meaning, it is also necessary to distinguish different types of semantic knowledge. Words do not only convey isolated concepts. They are shortcuts that allow us to activate a network of world knowledge, from objects to specific everyday experiences up to general representations of events, as well as their mutual interaction. 
Moreover, some aspects of semantics are associated with \textit{usage}: extra-linguistic,  pragmatic contexts in which language is learned and used contribute to shape a word meaning \citep{bolognesi2020words}.

From a linguistic perspective, \textit{Frame Semantics}  \citep{Fillmore77, Fillmore85, Fillmore09} proposes an approach implementing such information, based on different \textit{frames} representing the ``\textit{scenes}''. This notion, initially introduced in artificial intelligence  \citep{Minsky75}, presents not only the description of the meaning, but also how this meaning is realized in language, by specifying the linguistic characteristics of its constituents, including -- in some cases -- their lexicalization. In this section, we present the main features of Frame Semantics in relation with the question of accessing the meaning.
Generally speaking, frames can represent different types of knowledge, such as experiences, events, but also objects, etc. Frames condition the comprehension of linguistic expressions. They make the interpretation possible thanks to the description of a complete context. Describing a situation requires first to identify the different participants that will play a role in the semantic description. These participants correspond to semantic roles. Taking the classical example of a commercial transaction scenario, here are the different roles (called \emph{frame elements}) involved in this frame, as proposed in the FrameNet database \citep{Ruppenhofer10}:\footnote{https://framenet.icsi.berkeley.edu/} \emph{buyer, goods, money, seller} (these elements constitute the core of the semantic representation of the transaction, which can be completed with non-core elements such as \emph{means, rate, units}; see Table \ref{commercial1}).

At the lexical level, words convey meaning not only as isolated concepts, but also in connection with the frames they evoke. For example, the verbs \emph{sell} and \emph{buy} are related to the \textsc{commercial transaction} frame, the former with the perspective of the \textit{merchant}, the other from that of the \textit{customer}. In terms of linguistic structure, frames also implement the syntax/semantic relationship by providing interfaces between the semantic and syntactic roles (which is important when trying to explain non-compositional mechanisms). Going back to the \textsc{commercial transaction} frame, the verb ``\emph{to buy}'' has two obligatory arguments (\emph{buyer} and \emph{good}) and two optional ones (\emph{seller} and \emph{price}); conversely, the verb ``\emph{to buy}'' shares the same semantic slots, but in a different order (see Table \ref{commercial2}). In other words, these two verbs lexicalize different perspectives on the same knowledge. 

\begin{table}[h]
\small
\begin{center}
	\begin{tabular}{|l|l|l|l|l|}\hline
    	{\sc buyer} & V & {\sc goods}	& {\sc seller} & {\sc price} \\\hline
    	{\tt subject}	&& {\tt object} & {\tt from} & {\tt for} \\\hline
    	\emph{Angela}	& \emph{bought} & \emph{the book} & \emph{from Pete} & \emph{for 10\$}\\\hline
    	\emph{Eddy}	& \emph{bought} & \emph{them} &  & \emph{for 10\$}\\\hline
    	\emph{Penny}	& \emph{bought} & \emph{a bicycle} & \emph{from Stephen} & \\\hline
	\end{tabular}
\end{center}
\caption{Commercial transaction frame}	
\label{commercial1}
\end{table}

\begin{table}[h]
\small
\begin{center}
	\begin{tabular}{|l|l|l|l|l|l|}\hline
		{\sc verb} & {\sc buyer} & {\sc goods}	& {\sc seller} & {\sc money} & {\sc place} \\\hline
		\emph{buy} &	{\tt subject} & {\tt object} & {\tt from} & {\tt for} & {\tt at}\\\hline
		\emph{sell} &	{\tt to} & {\tt object} & {\tt subject} & {\tt for} & {\tt at}\\\hline
		\emph{cost} &	{\tt iobj} & {\tt subject} &  & {\tt object} & {\tt at}\\\hline
		\emph{spend} &	{\tt subject} & {\tt on} &  & {\tt object} & {\tt at}\\\hline
	\end{tabular}	
\end{center}
\caption{Commercial transaction scene}	
\label{commercial2}
\end{table}

Conversely, psycholinguistic evidence supports the idea that knowledge of real-world events is crucial in guiding online sentence processing. Generally speaking, the world knowledge that is produced and conveyed in language can be seen as a set of \emph{events} describing a scene or more generally the abstract knowledge of a situation. Events correspond to a general level of representation of meaning, not necessarily linked to a specific form, but implementing contextual information in a dynamic way \citep{Elman14}, during the interpretation of a given input. As described in \cite{McRae09}, people build and memorize general or prototypical forms of events rather than detailed descriptions, forming the \emph{Generalized Event Knowledge} (hereafter \emph{GEK}) stored in the long-term memory and encoding contextual information, coming from usage. 
\emph{GEK} is the abstract description of the knowledge of the world of an individual, which contains both entities and events. It is both theoretical and experience-based: some aspects of \emph{GEK} are cultural, common-sense, shared by a community, and some others are specific to the individual. While Frame Semantics and GEK have been developed in different fields, they both converge in the idea that every lexeme is interpreted against the background of a whole network of concepts in a particular domain \citep{Rambelli2024}.  


\section{Formalization: The Objects to be Processed}

In this section, we present the basis of a new approach integrating usage-based information into a clear syntax-semantic interface framework. 
The idea consists in gathering \emph{Distributional Semantics} into \emph{Construction Grammar}. This new approach, called \emph{Distributional Construction Grammar}, relies on a specification of three basic types of objects: {\bf constructions}, {\bf frames}, and {\bf events}, making it possible to describe respectively, syntax-semantics interface, background knowledge semantics, and usage-based semantics. 

\subsection{Constructions}


\emph{Construction Grammar} (hereafter \emph{CxG}; \citealp{Fillmore88,Goldberg03,HoffmanTrousdale2013,Hilpert2019,Hoffmann2022}) encompasses a variegate number of researches all taking a bottom-up rather than a top-down approach to language, and founded on the core notion of construction. 
In \emph{CxG}, a construction is a form/meaning pair in which the aspects of the meaning are not necessarily predictable from the form, implementing the non-compositional relation to the meaning. The two parts of a construction \emph{Cx} contain different kinds of information:  \textit{form} comprises phonological, morphological and syntactic structure; conversely, meaning includes semantic, pragmatic and discourse-functional meaning (cf. Figure~\ref{fig:cxg}). Accordingly, there are different shades of form and meaning, interconnected by symbolic links, which are typically arbitrary and established through convention \citep{hilpert2021}.

\begin{figure}
    \begin{center}
    \fbox{
    \begin{minipage}{5cm}
    
    \begin{minipage}{5cm}
    \fbox{
        \begin{avm}
        form \[
            \rm\it Syntactic properties  \\
            \rm\it Morphological properties  \\
            \rm\it Phonological properties \]
        \end{avm} }
    \end{minipage}

    \begin{minipage}{5cm}
        \fbox{
        \begin{avm}
        meaning \[
            \rm\it Semantic properties \\
            \rm\it Pragmatic properties \\
            \rm\it Discourse properties \]
        \end{avm}}
    \end{minipage}
    \end{minipage}
    }
    \end{center}
    \caption{Adapted from \citet{CroftCruse2004}.}\label{fig:cxg}
\end{figure}

\noindent{}

Among the several constructionist formalizations, it is possible to identify a group of frameworks closely related for their `formal' nature \citep{ungerer2023constructionist}: the \emph{Sign-Based CxG} \citep{Sag12}, the \textit{Fluid CxG} \citep{Steel2017}, and the \textit{Embodied CxG} \citep{BergenChang2005}. These approaches present the advantage of integrating the assumptions of CxG within a formalized framework. They all represent constructions in the form of feature structures, and more specifically, as attribute-value matrices (AVMs). Overall, they offer a way to define explicitly the relations between the different components of a construction. This aspect is of deep importance when gathering different sources of information, like in \emph{CxG}. 

In this section, we present our \emph{CxG} main features, following the formalization given in \emph{Sign-Based Construction Grammar} (hereafter \emph{SBCG}). We will adapt marginally this formalization in order to fit with the integration of the specific semantic representation that we are adopting (cf. next section).

A first remark about constructions concerns lexicalization. Differently from many other formalisms, there is no restriction about this feature: some \emph{Cx} can be attached to lexical entries (as illustrated in Figure \ref{AVM-frame1}), others can be more abstract, such as the covariational construction {\em The Xer the Yer} described above.  In this case, the form of the construction corresponds to a kind of morpho-syntactic pattern with which the realized input may match. As a consequence, a construction can describe linguistic objects of any type: word, phrase, pattern, sentence. This aspect is of deep importance in the \emph{activation-by-cues} mechanism that we propose for recognizing a construction, taking advantage of any type of feature, being it lexical, syntactic or from another domain.

Generally speaking, the role of a construction is to make explicit the relation between the form and the meaning thanks to a set of relations between the different features describing these two aspects (in particular by sharing values between different elements of the \emph{Cx}). 
In our representation, a construction is basically represented by the original {\sc form} and {\sc meaning} features.\footnote{Note that we use the label {\sc form} for the description of the morphological, phonological, syntactic features of the construction. This is a difference with SGBG which uses this label for the representation of morphological features only.}

\subsubsection{Form Features}
The {\sc form} feature contains the basic characteristics of constructions, potentially coming from the different linguistic domains not directly related to meaning. The following structure represents its organization:

    \begin{figure} 
        \begin{center}
            \begin{avm}
            	\[  form \[	phon	\< ... \>\\
            				surface\_form \< \rm\it f1, f2, ... fn \>\\
            				syn\[	cat \rm\it pos \\
            						val \< \it item-list\>\\
            						marking \rm\it marker\] \\
            				properties \[lin	\rm\it linearity  \\
            							 adj	\rm\it adjacency  \]\\
            	\]\]
            \end{avm}		
		\end{center}
        \caption{Sign-Based CxG  formalism: Schematic representation.}\label{fig:avm_scheme}
\end{figure}

\noindent{}The {\sc phon} feature (not detailed here) implements all aspects of the phonological form. Recall, as presented in \cite{Sag12}, that the content of this feature is a prototype, specifying the core phonological characteristics that can be adapted in more specific realizations. This principle applies to all feature values: variation or deviation from the prototype is always possible. This observation directly motivates our proposal to integrate distributional information into semantics using a vector-based representation (see Section 3.3).

We propose to represent the morphological aspects and more generally all surface forms (in the case of multiwords constructions) by a new feature called {\sc surface\_form} (corresponding to the feature {\sc form} in \emph{SBCG}). Basically, this feature contains the words (or set of words). 

In our approach, syntactic information is split in two: the feature {\sc syn} implementing the basic (morpho)syntactic characteristics, and the feature {\sc properties} representing explicitly the relations between the elements of the construction (e.g., linear order, cooccurrence, etc.). This type of information, usually not represented as such in \emph{SGBG} or other theories, is borrowed from the \emph{Property Grammars} framework \citep{Blache05a, blache2011evaluating,Blache16b}. 
The following list presents the set of features belonging to {\sc syn}:

	\begin{itemize}
        \item {\bf Category} ({\sc cat}): main characteristics of a construction
          \begin{itemize}
            \item {\sc type}: part-of-speech (V, N, etc.)
            \item {\sc case}: \emph{nominative, accusative}, etc.
            \item {\sc verb-form}: \emph{finite, infinitive, base, etc.}
            \item {\sc aux}: auxiliary (boolean)
            \item {\sc inv}: subject-verb inversion (boolean)
            \item {\sc ic}: independent clause (boolean)
            \item {\sc select}: elements selected by another element, typically modifiers (e.g., adjectives, adverbs) or markers (e.g., determiners). They have to be distinguished from complements (mentioned in the valence list)
            \item {\sc xarg}: external non-local argument, outside from the maximal projection. For example, long-distance dependencies are external arguments.
             
          \end{itemize}
        
        \item {\bf Valence} ({\sc val}): set of elements subcategorized by the sign under description.
		\item {\bf Mark} ({\sc marking}): constructions ``\emph{marked}'' with a marker.
		 For example: \emph{than} (compared phrase), \emph{as} (equated phrases, e.g., ``\emph{as I could}''), \emph{of} (some of-phrases, e.g., ``\emph{of mine}''), \emph{det} (determined signs of category noun), \emph{a} (a subtype of \emph{det}), \emph{def} (definite signs). 
            
    \end{itemize}

The {\sc properties} feature encodes syntactic relations (e.g., linearity, exclusion, adjacency, cooccurrence, etc.) as proposed in \emph{Property Grammars}. The role of such a feature is to encode explicitly any type of information making it possible to characterize and then recognize a construction. For example, the ditransitive construction (e.g., ``\emph{Mary gives John a book}'') is characterized by the fact that the verb governs three \emph{NP} arguments (\emph{Mary}, \emph{John} and \emph{a book}) that need to be in this specific order: the subject preceding the verb, in turn preceding the indirect object and the object.\footnote{At the difference with \emph{SBCG}, we encode  explicitly  in \emph{DCxG} linearity thanks to a specific feature.} Moreover, the indirect object and the object needs to be adjacent. These two characteristics are encoded in specific {\sc linearity} and {\sc adjacency} properties, as illustrated in Figure \ref{ditrans-prop}. 

\begin{figure}
	\begin{center}
        \begin{avm}
        	\[form
        	\[syn \[cat & \@1 V \] \\
			  properties \[lin \{\@2 NP$_{[subj]}$ $\prec$ \@1 $\prec$ \@3NP$_{[obl]}$ $\prec$ \@4NP$_{[obj]}$\} \\
			  		 adj \{\@1 $\oplus$ \@3; \@3 $\oplus$ \@4\} \]					
			\]\]
        \end{avm}						
	\end{center}
\label{ditrans-prop}
\caption{Properties of the ditransitive construction}	
\end{figure}

Note that several studies on the ditransitive construction reveal additional types of constraints. For instance, \cite{Biber99}, in a corpus-based study, confirm a preference for linearizing light constituents before heavy ones. This helps explain why, in double object constructions, recipients are most often realized as single words. Such preferences can be directly encoded in the form description by adding specific constraints to the construction.

In some cases, it may be necessary to completely override the default construction. For example, \cite{Siewierska07} describe non-canonical patterns in the ditransitive construction in the Lancashire dialect, where personal pronouns in non-subject argument positions lead to atypical word orders. This implies that, depending on the morphological form of the argument, the standard linearization constraint may be altered.

Such phenomena can be captured through constraint relaxation (see Section 3.3) or by modifying constraint weights (see Section 4.1), allowing for flexible adaptation of the construction to specific contexts.

\subsubsection{Meaning Features}

The {\sc meaning} features contain the intrinsic semantic characteristics of the construction (feature {\sc sem}), the entity that is possibly involved by the construction (typically the referential object when the construction corresponds to a noun) and the event (or set of events) it activates (feature {\sc events}).\footnote{These two last features will be described in detail in the next section.}  The general representation corresponds to the following structure:
		\begin{center}
            \begin{avm}
            	\[meaning
            	\[ 	sem \rm\it semantic features \\
					entity \rm\it sign \\
					events \< \rm\it event-list\> \]\]
			\end{avm}		
		\end{center}						

	\begin{itemize}
		\item  {\sc sem}: intrinsic characteristics, description of the specific attributes of the construction: semantic features, aspect, attributes, properties, etc. This structure contains three features (cf. Figure~\ref{fig:book_eat_ex}):
		\begin{itemize}
			\item {\sc index}: the index of the semantic structure (the label), specifying a situation, an entity described in a frame

			\item {\sc frames}: contains the different elements composing the frame
			\item {\sc ltop}: indicates the top frame in the frame hierarchy (corresponding to the complete interpretation in the case of a sentence)
		\end{itemize}
\begin{figure}
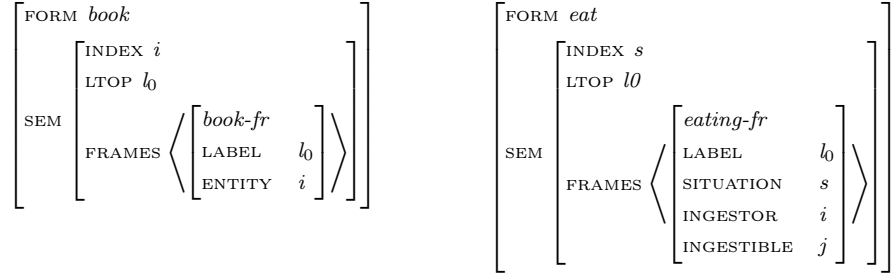


			\begin{center}
	        \begin{avm}
            \[form \rm\it book \\
            	sem \[
                        index  \rm\it i \\
              			ltop  \rm\it l$_{0}$ \\
              			frames \<\[\rm\it book-fr \\
              					label & \rm\it l$_{0}$ \\
								entity & \rm\it i\]\>\]\]
			\end{avm}
			\hspace{1cm}
			\begin{avm}
            \[form \rm\it eat \\
            	sem \[index  \rm\it s \\
              		  	ltop  \rm\it l0 \\
              			frames \<\[\rm\it eating-fr \\
              				label & \rm\it l$_{0}$ \\
							situation & \rm\it s\\
							ingestor & \rm\it i \\
							ingestible & \rm\it j\]\>\]\]
			\end{avm}		
		
			\end{center}
    \caption{Example of the {\sc sem} component for the noun \textit{book} and the verb \textit{to eat}.}
    \label{fig:book_eat_ex}
\end{figure}

		\item {\sc entity}: an entity is a referential object, defined by a set of attributes representing its properties. It has a conceptual 
reality, a self-contained existence
        \begin{center}
        	\begin{avm}
        	\[	\it entity\_name \\
        		attributes \<\it a1, a2, ..., an\>\]
        	\end{avm}
        \end{center}

		\item  {\sc event}: specifies the event or the set of events that are related to the construction. For example, in the case of a verb, the activated event corresponds to the thematic relations. This feature is described in detail in the next section.
	\end{itemize}

\subsubsection{Argument Structure}

The argument structure encodes the combinatorial organization of a lexical sign, by specifying the subcategorization lists of the categories it governs. The following table, borrowed from \citep{Sag12}, illustrates this information associated with different types of verbs:

	\begin{center}
    	\begin{tabular}{lll}
    		<NP> & \emph{die, laugh}, $\ldots$ & (intransitive) \\
    		<NP, NP> & \emph{hit, like}, $\ldots$ & (transitive) \\
    		<NP, NP, NP> & \emph{give, tell}, $\ldots$ & (ditransitive) \\
    		<NP, NP, PP> & \emph{put, place}, $\ldots$ & (indirect transitive) 
    	\end{tabular}
    \end{center}
    
It is important to note that the feature {\sc arg-st} is only relevant (or appropriate in the \emph{SBCG} terminology) for lexical signs.  \emph{SBCG} distinguishes between valence and argument structure: the argument structure encodes overt and covert arguments, including extracted (non-local) and unexpressed elements, unlike the valence feature {\sc val} in the form description, which represents only realized elements. When no covert arguments occur, the {\sc arg-st} and {\sc val} features are identical.

The {\sc arg-st}  implements the interface between syntactic and semantic roles. The arguments are ordered by their accessibility hierarchy (\emph{subj} $\prec$ \emph{d-obj} $\prec$ \emph{obl} ...), encoding the syntactic role. Each argument specifies the case, related to the grammatical function and the thematic roles.

\subsubsection{Complete Constructions}

In \emph{SBCG}, linguistic information is encoded in the grammar by means of \textit{attribute-values structures} (AVS) that describe lexical entries as well as abstract constructions. In turn, a construction is based on the description of its participants and their relations. Building the linguistic description of an input (for example a sentence) consists in instantiating and aggregating the different relevant constructions. Classically, this process is based on a unification operation. After scanning an input word, the parsing mechanism consists in retrieving its corresponding description in the lexicon (an AVS) and scanning the different activated constructions describing the current input in order to check whether any expected argument (in the {\sc form}  feature) from one of these constructions can be unified with it\footnote{Our framework relies on the SBCG formalism, but other approaches have been proposed recently to represent how constructions are combined/blended in
the working memory, such as CASA \citep{herbst2018construction}.}. 

\begin{figure}
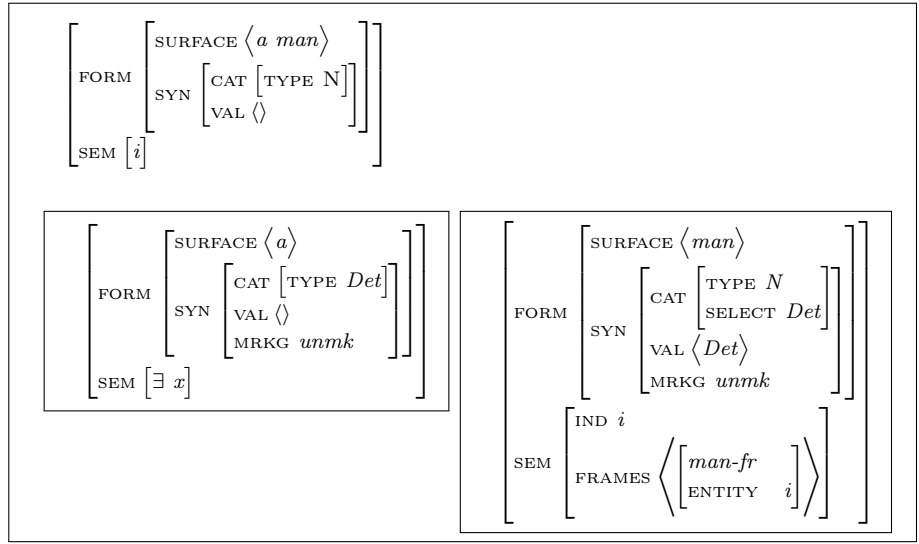

	\begin{center}
		\fbox{
		\begin{tabular}{l}
			\begin{avm}
				\[form \[
					surface \<\rm\it {a man}\> \\
                  	syn \[ cat \[ type N \] \\
                           val \< \> \] \]\\
                  sem \[ \rm\it i\] \]
				\end{avm}
			\\ \\
			\fbox{
                \begin{avm}
					\[
					form \[
						surface \<\rm\it {a}\> \\
						syn \[ cat \[type \rm\it Det \] \\
                    		val \<\> \\
    						mrkg \rm\it unmk \] \]\\
					sem \[ $\exists$ \rm\it x \]  \\
					\]
                \end{avm}
                }
			\fbox{
                \begin{avm}
                \[  form \[
                	surface \<\rm\it {man}\> \\
                	syn \[ cat \[type \rm\it  N  \\
        						select \rm\it Det\] \\
        				  val \<\rm\it Det\> \\
        				  mrkg \rm\it unmk \] \]\\
                   sem \[ ind \rm\it i \\
                   			frames \< \[\rm\it man-fr \\
								entity & \rm\it i\]\> \]  \]
                \end{avm}
                }
		\end{tabular}}
	\end{center}
\caption{The structure of the phrase \emph{a man}}   
\label{example-cx1} 
\end{figure}

A simplified representation of constructions can be done with embedded boxes, distinguishing between the information related to the entire construction and that of its participants. The  example  in Figure \ref{example-cx1} presents the construction corresponding to the NP `\emph{`a man}''. In particular, it illustrates the selection/subcategorization of the determiner by the noun via the {\sc select} and {\sc val} features.


\begin{figure}
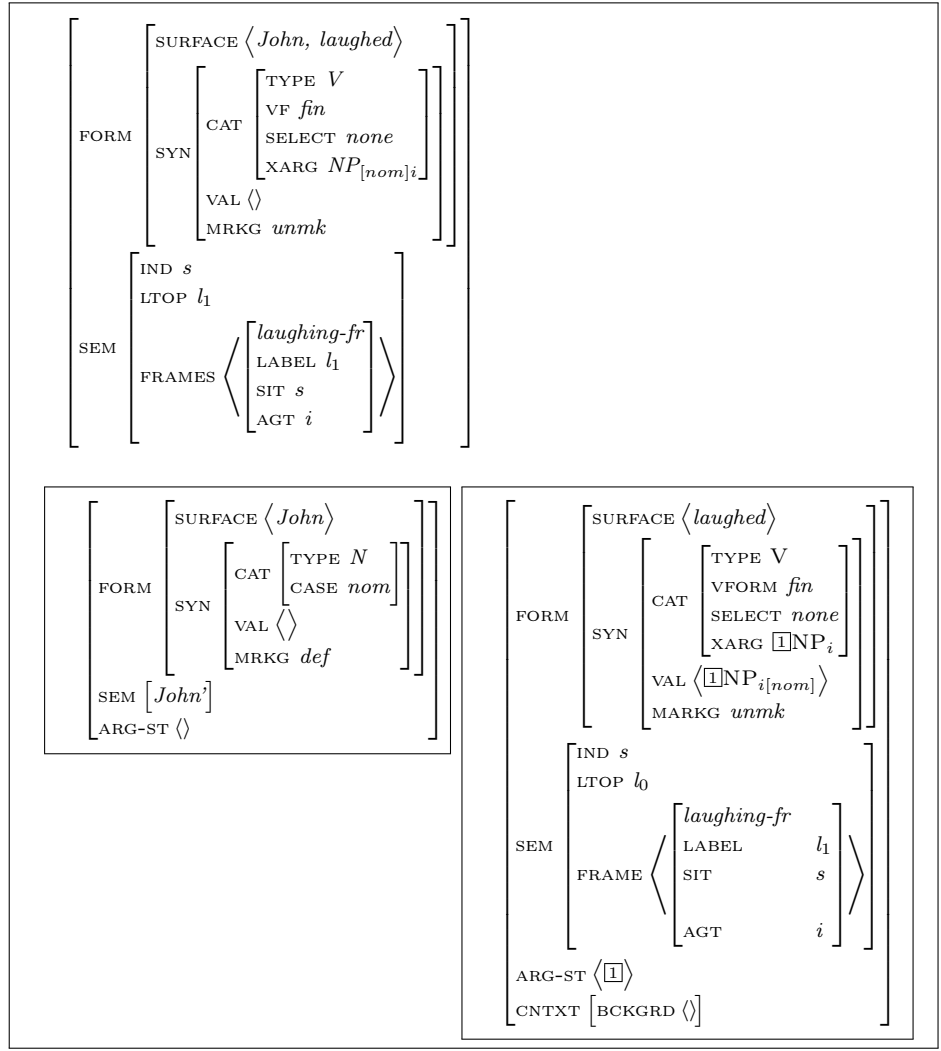

	\begin{center}
    \fbox{
    \begin{tabular}{l}
        \begin{avm}
        \[  form \[
        	surface \< \rm\it John, laughed \>\\
			syn\[	cat \[ type \rm\it V \\
						vf \rm\it fin \\
						select \rm\it none \\ 
						xarg \rm\it NP$_{[nom]i}$ \] \\
					val \<\> \\
					mrkg \rm\it unmk\]\]\\
			sem \[  ind \rm\it s \\
					ltop \rm\it l$_{1}$ \\
					frames \< \[\rm\it laughing-fr \\
								label \rm\it l$_{1}$ \\
								sit \rm\it s \\
								agt \rm\it i \]\> 
				\]
		\]
		\end{avm}\\\\
		\fbox{
		        \begin{avm}
        \[ form \[
        	surface \<\rm\it {John}\> \\
           	syn \[ cat \[type \rm\it  N  \\
                    	case \rm\it nom \] \\
				  val \<\rm\it \> \\
				  mrkg \rm\it def \]\]\\
           sem \[ \rm\it John' \] \\
           arg-st \< \>\\
           \]
        \end{avm}}
        \fbox{
        \begin{avm}
            \[form \[
            	surface \<\rm\it {laughed}\> \\
              	syn \[ cat \[	type   V  \\
              				vform \rm\it fin \\
                        	select \rm\it none \\
							xarg \@1NP$_{i}$ \]\\
					val \< \@1NP$_{i[nom]}$ \> \\		
					markg \rm\it unmk \]\]\\
              sem \[ ind  \rm\it s \\
              		ltop \rm\it l$_{0}$ \\
					frame \<\[\rm\it laughing-fr \\
							label & \rm\it l$_{1}$ \\
							sit & \rm\it s \\ \\
							agt & \rm\it i\]\> \]\\
              arg-st \<\@1\> \\
			  cntxt \[bckgrd \<\>\]
			  \]
        \end{avm}
        }
		\end{tabular}}
	\end{center}
\caption{The phrase \emph{John laughed}}   
\label{example-cx2} 
\end{figure}

At the sentence level, the same type of representation can be given, as illustrated in Figure \ref{example-cx2}. This construction illustrates different aspects of \textit{structure sharing}. For example, the  valence list of the verb \emph{laughed} is an NP of index \fbox{\footnotesize 1}. This same index is found as value of the argument structure list of the same verb as well as its external argument, indicating the same element will be instantiated there. The same mechanism occurs for propagating the semantic values from the participant \emph{laughed} to the root of the construction.

Abstract constructions that do not involve lexical material can be represented in the same way. The  structure in Figure \ref{abstract-cx}.a represents the \emph{subject-predicate construction}, stipulating that the verbal sign selects the subject via the {\sc val} feature: the subject, indexed by \fbox{\footnotesize 3}, is shared as a value in the valence feature of the verb. In the same way, the categorical information of the verb is propagated to the constructions characterization thanks to the same structure sharing mechanism. The ditransitive construction depicted in Figure \ref{abstract-cx}.b includes the semantic aspects only, without details about the form and the participants. The main semantic feature of this construction is that whatever the lexicalization of the verb, it always involves a possession interpretation (more precisely the transfer of something to somebody), implemented in the \emph{transfer} frame.

\begin{figure}
\begin{tabular}{>{\centering\arraybackslash}m{.5\textwidth} >{\centering\arraybackslash}m{.4\textwidth}}
    \fbox{
        \begin{tabular}{l}
        \begin{avm}
        	\[	syn \[ cat \@1 \\
					 	val \<\> \] \\
			\]
        \end{avm}

        \\ \\
    
        \fbox{
        	\begin{avm}
        	\@3\[
        	syn \[\[gf \rm\it subj \] \] \]
        	\end{avm}
        }
        \fbox{
        	\begin{avm}
 			\[syn \[cat \@1\[vf \rm\it fin \\
        				inv -
						aux -\] \\
			mark \rm\it unmk \\
			val \<\@3\>\]  \]
        	\end{avm}
        }
        \end{tabular}}
        &
	\fbox{
		\begin{avm}
			\[  syn \[cat \rm\it V\]\\
				arg-st \< NP$_{x}$, NP$_{y}$, NP$_{z}$ \> \\
				sem \[ frames \<\[ \asort{transfer-fr} 
									agent &\rm\it x \\	
									recipient &\rm\it y \\	
									theme &\rm\it z \]\>\]
			\]
		\end{avm}}\\

		Figure \ref{abstract-cx}.a: Subject-predicate construction & Figure \ref{abstract-cx}.b: Ditransitive construction \\
\end{tabular}
\label{abstract-cx}
\end{figure}

\subsection{Frames and Events}

In the following, we adopt an attribute-value matrix representation \citep{Sag12}. In this representation, each information corresponds to a feature whose values can be of different types. In the next section, we describe more precisely the different aspects of this type of representation, concerning in particular the structure sharing mechanism, represented by structure co-indexation. As illustrated in Figure \ref{AVM-frame1}, this structure encodes the main types of information concerning syntax, semantics and their interface through the argument structure.

\subsubsection{Representation of  Frames}
As introduced in Section 2.3, 
Frame Semantics proposes an approach to implement such information, based on different frames representing the ``scenes". Frame Semantics and Construction Grammar are historically linked; \citet{michaelis2013} already presented a way to include frames into SBCxG representation. Following that work, 
Figure \ref{AVM-frame1} illustrates the encoding of the different types of information associated with the verb \emph{buy}.

\begin{figure}[h]
\begin{center}
	\begin{avm}
            \[	phon \<\rm\em {buy}\> \\
                syn \[ cat \rm\em  V  \\
                       val \<\@1NP$_{[subj]}$, \@2NP$_{[obj]}$, \@3PP$_{[obl]}$,\> \] \\
                sem \[\rm\em commercial\_transaction\_fr \\
                	frame\_element  \@1 \[\rm\em buyer\] \\
                	frame\_element  \@2 \[\rm\em goods\] \\
                	frame\_element  \@3 \[\rm\em seller\] \\
                	frame\_element  \@4 \[\rm\em money\] \\
					\] \\
            	arg-st \< 
            	\[ 	syn  \| cat \rm\em N  \\
					sem 	\@1\],
            	\[ 	syn  \| cat \rm\em Prep \\
					sem 	\@2\], 
            	\[ 	syn  \| cat \rm\em N   \\
					sem 	\@3\],  
            	\[ 	syn  \| cat \rm\em Prep \\
					sem 	\@4\]\> 
			\]
	\end{avm}
\end{center}
\caption{Attribute-value representation of \emph{buy}}	
\label{AVM-frame1}

\end{figure}

The presented example proposes a frame linked to a specific lexical entry. However, we could have frames linked to entire constructions. This idea aligns with Construction Semantics (CxS; \citealt{willich2022introducing}), which models the semantic properties associated with constructions in terms of three types of frames: \textit{lexical} (a frame evoked by a lexical unit), \textit{constructional} (a frame evoked by the construction as a whole), and \textit{construct frame} (result of combining lexical and constructional frames for a specific construct). While our proposal implicitly considers these different layers of analysis, we do not provide a specific distinction as CxS, but rely on a unique representation.

\subsubsection{Representation of  Events}
In our approach, generic semantic information is encoded by {\bf frames} representing the different semantic roles and their characteristics, and specifying their relation with other frames (e.g., inheritance, precedence, perspective, etc.). On their side, {\bf constructions} implement the form/meaning interface by specifying on the one hand the participants and on the other hand their relations with the meaning. Constructions, as frames do, also implement generic information, independent from the specific context of the input. In such an organization, the role of {\bf events} is to bridge the gap between generic and contextual semantic information by specializing the meaning encoded in constructions. For example, if the word \emph{student} appears in the context of the verb \emph{read} as subject, then the most prototypical object becomes \emph{book} \citep{lenci2011composing}. Events are then considered as functions that specialize the semantic meaning by adding new relations or new values. In the same example, the  \emph{student-reading} event specifies the argument structure by specifying the semantic prototype of the object (in this case, \emph{book}).

In our proposal, we consider events as a way to integrate contextual and distributional information into the frames. More precisely, an event brings specific information making it possible to precise or refine the generic semantic description given by constructions. In other words, the {\bf frames} define a \emph{prototypical} semantic representation based on the different semantic roles (the frame elements), and the {\bf events} provide a \emph{specialization} of the frame by taking into account the context and specifying the participants and relation between them. For example, the frame associated to the action of \emph{reading activity} is described in \emph{FrameNet} by the \textit{reading\_fr} frame as follows:

	\begin{center}
		\begin{avm} \[\it reading\_fr \\
					reader  \rm\it person\\
					text  \rm\it entity\\
					topic  \rm\it idea \]
		\end{avm}
	\end{center}

\noindent{}Taking again the scenario of a student reading, the prototypical object becomes the entity \emph{book} (in other words, students typically read books). This specialization of the frame \emph{reading\_fr} can then be implemented by the event \emph{student\_read\_event} as represented in Figure \ref{event1}. Note that we introduce in this structure a new feature, {\sc specialize},  which is specific to event descriptions and specifies the construction to be completed. 

\begin{figure}
	\begin{center}
        \avmsortfont{\it}
        \avmvalfont{\it}
        \begin{avm}
        \[\asort{student-read-event} 
        	specialize	& \rm\it read-lexeme-cx \\
             	sem & \[ frame\[\asort{reading-fr} 
            			reader  &\rm\it \@1 \\
            			text	 &\rm\it \@2 
        				\]\] \\
           	arg-st & \< 
            		\@1\[syn &\[ cat \[ type & N \\
            						  case & \rm\it nom\] \\
            				   val \<\> \]\\
            			sem  &\[ind  \rm\it i \\
            					frames \<\[\asort{student-fr}
            								entity & i \]\>\]\],
            		\@2\[syn &\[ cat \[ type & N \\
            						  case & \rm\it acc\] \\
            				   val \<\> \]\\
            			sem  &\[ind  \rm\it j \\
            					frames \<\[\asort{book-fr}
            								entity & j \]\>\]\]
            			\>\\
        		\]
        \end{avm}
	\end{center}
\caption{Description of the \emph{student-read} event}
\label{event1}
\end{figure}

As can be seen in this example, an event describes the specific situation in which one or several participants of the action (represented in the frame) involve specific relation between arguments. An event is then a specific kind of construction implementing context-dependent information. Differently from constructions, events are not associated with a form and only focus on the semantics of the structure. In other words, events disentangle the conceptualized meaning from its concrete form. We distinguish then events from constructions, the former representing the meaning, the latter the condition of its realization in language. 

To summarize, we propose in our approach a model of sentence comprehension based on three basic components: 




\begin{itemize}
	\item {\bf Frames}: represent the generic semantic knowledge of the word, describing scenes, situations
	\item {\bf Constructions}: represent the linguistic forms, describing their interpretation thanks to the form/meaning pairings, the meaning being encoded by frames
	\item {\bf Events}:  represent the dynamic (or contextual) part of the semantic information by specializing the relations between the different elements of a frame via the specification of construction participants
\end{itemize}	

When adopting a more general view in the perspective of linguistic theory, this distinction fulfills the requirements of a theory in which all different sources of information, coming from the different linguistic domains (morphology, phonology, syntax, semantics, pragmatics, etc.), are at the same level. In other words, no domain is a pre-requisite for the description of another one. This is a major step towards a theory that is not syntactico-centered (which is still the case of a majority of linguistic theories), and that would not require to consider the syntactic structure as a pre-requisite for building the semantic representation. In \emph{DCxG}, the different sources of information are encoded separately in these three components, that are interacting (by sharing information) but can still be instantiated separately. Moreover, each component is not necessarily complete in the sense that it does not have to cover the entire input.\footnote{In a graph-based approach in which linguistic information is represented in terms of graphs (among which, trees), this comes to say that the structure is not necessarily connected.} 

Therefore, a {\bf grammar in \emph{DCxG} is the set of the three components: frames, constructions and events}. Accessing these components can be done to several manners, in particular through similarity or analogical reasoning \citep{Rambelli22, Rambelli2024}.


\subsubsection{Examples}
The example of the \emph{student-read} event illustrates this organization of linguistic knowledge. Table~\ref{tab:student_read_ex} illustrates the different pieces of information when processing the partial input ``\emph{Students read} ...''.
\begin{table}[htb]
\begin{center}
	\begin{tabular}{|l|l|}\hline
		{\sf Frame}	&	\emph{reading-fr} \\
				&	\emph{student-fr} \\
				&	\emph{book-fr}	\\\hline
		{\sf Constructions} & \emph{subject-predicate-cx} \\
					& \emph{student-lexeme-cx} \\
					& \emph{read-lexeme-cx} \\\hline
		{\sf Event}	& \emph{student-read-event} \\\hline
	\end{tabular}	
    \caption{Information activated when processing the partial input ``Students read ...''}\label{tab:student_read_ex}
\end{center}
\end{table}

\noindent{}At the initial stage, the available information consists of the different lexical frames and constructions. When scanning the beginning of the input sentence, the lexical constructions corresponding to \emph{students} and \emph{read} are activated. When both of them become accessible, the conditions for activating \emph{student-read-event} are filled. As a result, the new construction describing the partial sentence is instantiated and the object corresponding to \emph{book} is pre-filled.

\begin{figure}
	\begin{center}
    \framebox[11.5cm]{
	\hspace{-.7cm}
    \begin{tabular}{m{5cm} m{5cm}}
        \begin{avm}
        \[  form \[
        	surface \< \rm\it students, read \>\\
			syn\[	cat \[ type \rm\it V \\
						vf \rm\it fin \] \\
					val \<\> \]\]\\
			sem \[ frames \< \[\rm\it reading-fr \\
								reader  &\rm\it \@1 \\
            					text	 &\rm\it \@2 \]\> 
				\]
		\]
		\end{avm}\\\\
		\begin{minipage}{4cm}
		\fbox{
		\begin{avm}
        \@1\[ form \[
        	surface \<\rm\it {students}\> \\
           	syn \[ cat \[type \rm\it  N  \\
                    	case \rm\it nom \] \\
				  val \< \> \]\]\\
           sem \[ind  \rm\it i \\
            	 frames \<\[\asort{student-fr}
            			entity \rm\it  i \]\>\] 
           \]
        \end{avm}}
        
		\fbox{
        \begin{avm}
        \@2\[ form \[
        	surface \<\rm\it {book}\> \\
           	syn \[ cat \[type \rm\it  N  \\
                    	case \rm\it acc \] \\
				  val \< \>  \]\]\\
           sem \[ind  \rm\it j \\
            	frames \<\[\rm\it {book-fr}\\
            	entity \rm\it  j \]\>\] 
           \]
        \end{avm}}
        \end{minipage}
		&    
        \fbox{
        \begin{avm}
            \[form \[surface \<\rm\it {read}\> \\
              		 syn \[ cat \[	type   V  \\
              				vform \rm\it fin  \]\\
					 val \< \@1NP$_{i}$, \@2NP$_{j}$ \>  \]\]\\
              sem \[frame \<\[\rm\it reading-fr \\
							reader  \rm\it i \\ 
							text \rm\it j\]\> \]\\
           	  arg-st  \< 
            		\@1, \@2 
             \>
			 \]
        \end{avm}
		}
		\end{tabular}}
	\end{center}
\caption{The partial sentence \emph{Students read ...}}   
\label{example-cx4} 
\end{figure}

The second example we propose, based on the \emph{ditransitive construction}, shows how events can instantiate an abstract construction dynamically, according to the context (entering then into distributional information). As shown above,  the ditransitive construction is of the following form: \emph{NP$_{x}$ V NP$_{y}$ NP$_{z}$} (V being a verb, selecting the different NPs a complement). When focusing on NP$_{y}$ (the {\sc recipient} in the frame), let's imagine that we obtain the following completion (e.g., from a \textit{cloze} task): 

\begin{enumerate}
	\item \emph{NP$_{x}$  give students ...} $\rightarrow$ \emph{NP$_{x}$ give students} \underline{exercises} 
	\item \emph{NP$_{x}$  give children ...} $\rightarrow$ \emph{NP$_{x}$ give children} \underline{sweets} 
\end{enumerate}	

\noindent{}Then, we have a distributional restriction that can be encoded as an event stipulating the cooccurrences \emph{student/exercises} and \emph{children/sweets} an shown in  Figure \ref{ditrans-event}.

\begin{figure}
\begin{minipage}{15cm}
\begin{center}
	\begin{avm}
		\[  specialize \rm\it ditransitive-cx\\
			arg-st \< NP$_{x}$, NP$_{y}$, NP$_{z}$ \> \\
			sem \[ frames \<\[ \asort{transfer-fr} 
								agent &\rm\it x \\	
								recipient &\rm\it y$_{[\mathit{children}]}$ \\	
								theme &\rm\it z$_{[\mathit{sweets}]}$ \]\>\]
			\]
	\end{avm}
	\hspace{1cm}
	\begin{avm}
		\[  specialize \rm\it ditransitive-cx\\
			arg-st \< NP$_{x}$, NP$_{y}$, NP$_{z}$ \> \\
			sem \[ frames \<\[ \asort{transfer-fr} 
								agent &\rm\it x \\	
								recipient &\rm\it y$_{[\mathit{students}]}$ \\	
								theme &\rm\it z$_{[\mathit{exercises}]}$ \]\>\]
			\]
	\end{avm}
\end{center}
\end{minipage}
\caption{Ditransitive event specialization}
\label{ditrans-event}
\end{figure}

The same type of specialization can be applied when encountering some specific lexical units, restricting implicitly the context. For example the word ``\emph{shopkeeper}'' is implicitly associated to a kind of corner shop, entailing the specialization of the \emph{commercial-transaction} frame: in this case, there is an expected price, usually limited to a certain amount, the payment is done immediately by cash (in which case some change is possibly returned) or credit card, etc. This specialization is implemented by the \emph{corner-shop-event}, as illustrated in Figure \ref{corner-shop-ev}.

\begin{figure}
\begin{center}
	\begin{avm}
		\[\rm\it corner-shop-event \\
		 specialize \rm\it commercial-transaction-fr \\
		 sem \[ frame-element \[ 
		 			price \rm\it price\\
					payment  \{\rm\it cash, credit\_card\} \\
					amount\_given \rm\it money \\
					change \rm\it money \] \]
		\]
	\end{avm}
\end{center}
\caption{\emph{Corner-shop} event}
\label{corner-shop-ev}
\end{figure}

\subsection{Bringing Vectors into Constructions}

Our proposal consists in bringing together two theoretical paradigms, representing both distributional and formal grammatical knowledge: the idea is to integrate distributional information into constructions by completing the semantic structure of a lexical item with its distributional vector. This new information, embedded into the attribute-value structure, enriches the way to activate or instantiate the structure. 

As sketched above, the language processing mechanism in feature-based approaches such as \emph{SBCG} consists in \textit{unifying} the descriptions (also considered as \textit{set of constraints}) contained in the grammar with the signs progressively instantiated during the parsing of the input. Typically, scanning a new word consists in accessing its complete structure stored in the lexicon and unifying it with the expected elements. For example, a noun can be a complement of a verb, provided that it matches with the description (the expected properties) specified in the valence list of the verb. Even though in some cases the constraints of the description can be relaxed, unification remains a strict control in the sense that it fails or succeeds. 

Distributional semantics associates to lexical signs a vector representing its distribution in context. Comparing the \textit{similarity} between two vectors is straightforward (e.g., with the cosine) and makes it possible to evaluate directly the semantic similarity between two lexical signs (according to the distributional hypothesis, linguistic expressions appearing in similar contexts have similar meanings). Adding distributional vectors to lexical descriptions offers thus a new way, on top of unification, to match two structures: \textit{two lexical structures are compatible when they reach a certain similarity threshold}. 
The integration of this new information to the feature structure consists in adding a new feature, called {\sc ds-vector}, to the semantic features, as illustrated in Figure \ref{dsVect1}.

\begin{figure}
	\begin{center}
        \begin{avm}
        \[  form \rm\it {student} \\
        	syn \[ cat \[type & N  \\
                    	case &\rm\it c \\
						select &\rm\it Det\] \\
				  val \<\rm\it Det\> \\
				  mrkg \rm\it unmk \]\\
           sem \[ ind \rm\it i \\
           		  frames \<\[\asort{student-fr}
								entity & \rm\it i \]\> \\
				  ds-vector  $\overrightarrow{\mathit{student}}$\] \]
        \end{avm}
	\end{center}
\label{dsVect1}
\caption{Description of the lexeme \emph{student} with its distributional vector}
\end{figure}

It is interesting to note that distributional vectors can also be calculated by taking into account different features instead of lexical form only. Several works in distributional semantics have used in particular the notion of \textit{thematic fit} \citep{lenci2011composing,Sayeed16,Santus17,chersoni2017structure,chersoni2020word,marton2021thematic}. The idea consists in defining the most prototypical filler for a thematic role of a given verb. The structure in figure \ref{ds-vect3} presents the lexical description of the verb \textit{``read"}, in which different distributional vectors have been added: $\overrightarrow{\mathit{read}}$ that describes the distribution of the lexical entry itself, plus the vectors corresponding to the thematic roles specified in the \emph{reading} frame, namely $\overrightarrow{\mathit{reader}}$ and $\overrightarrow{\mathit{theme}}$. This information, when matched with the syntactic information (via structure sharing between {\sc val} and {\sc arg-st} feature values in the lexical construction depicted in Figure \ref{ds-vect3}), implements the specification of the arguments' typical roles (in this case, the typical reading subject and the typical reading object). 

\begin{figure}
        \begin{center}
            \avmsortfont{\it}
            \avmvalfont{\it}
            \begin{avm}
            \[\rm\it{read-lxm-cx}\\
            	form \[surface \<\rm\it {read}\> \\
            		syn \[ cat \[type & V  \\
            				 vform &\rm\it fin\] \\
            		    val \<\rm\it \@1, \@2\> \\
            			mrkg \rm\it unmk \]\]\\
            	arg-st  \< 
            		\@1\[syn & \[ cat \[ type & N \\
            						  case & \rm\it nom\] \\
            				   val \<\> \]\\
            			sem  & \[ind  \rm\it i \\
            					frames \<\[\asort{reader-fr}
            								entity & i \]\>\] \\
						ds-vector &  $\overrightarrow{\mathit{reader}}$\],
            		\@2\[syn & \[ cat \[ type & N \\
            						  case & \rm\it acc\] \\
            				   val \<\> \]\\
            			sem  & \[ind  \rm\it j \\
            					frames \<\[\asort{theme-fr}
            								entity & i \]\>\]\\
            			ds-vector &  $\overrightarrow{\mathit{theme}}$\]
            			\>\\
            	sem  \< \[\asort{reading-fr} 
            			reader  &\rm\it \@1 \\
            			theme	 &\rm\it \@2  \]\>\\
				ds-vector  $\overrightarrow{\mathit{read}}$\\
            \]
            \end{avm}   
        \end{center}
\caption{The \emph{read} lexeme construction: specification of thematic role}
\label{ds-vect3}
\end{figure}

Integrating a lexical construction into a structure during sentence processing consists in evaluating first the similarity between the different distributional vectors of the lexical entry (in Figure \ref{ds-vect3} those represented at the root and the argument structure), and then applying unification. This approach makes it possible to implement a new mechanism: \textit{loose unification}, relaxing when necessary unification failures.

Like constructions, events can also integrate distributional information by adding distributional vectors completing the description of the construction specialized by the event. The example of the \emph{student-read event}, described in Figure \ref{student-read-final}, specializes  the \emph{read-lexeme} construction by specifying the semantic value of its arguments. In the following, we use a compact representation of such structures by indicating as subscript the syntactic and semantic characteristics of the sign. For example, the arguments of \emph{student-read} event can be compacted by representing as subscript the syntactic function and the distributional vectors as presented in Figure \ref{compact-student-read}.  

\begin{figure}
	\begin{center}
        \avmsortfont{\it}
        \avmvalfont{\it}
        \begin{avm}
        \[\asort{student-read-event} 
        	specialize	& \rm\it read-lxm-cx \\
             	sem & \[ frame\[\asort{reading-fr} 
            			reader  &\rm\it \@1 \\
            			text	 &\rm\it \@2 
        				\]\] \\
           	arg-st & \< 
            		\@1\[syn &\[ cat \[ type & N \\
            						  case & \rm\it nom\] 
						  \]\\
            			sem  &\[ind  \rm\it i \\
            					frames \<\[\asort{student-fr}
            								entity & i \]\>\\
						ds-vector $\overrightarrow{\mathit{student}}$\]\],
            		\@2\[syn &\[ cat \[ type & N \\
            						  case & \rm\it acc\] \]\\
            			sem  &\[ind  \rm\it j \\
            					frames \<\[\asort{book-fr}
            								entity & j \]\>\\
						ds-vector $\overrightarrow{\mathit{book}}$\]\]
            			\>\\
        		\]
        \end{avm}
	\end{center}
\label{student-read-final}
\caption{Final representation of the \emph{student-read} event}	
\end{figure}

\begin{figure}
	\begin{center}
		\avmsortfont{\it}
		\avmvalfont{\it}
		\begin{avm}
			\[	form \<\rm\it {read}\> \\
				syn \[ 	cat \[type & V  \\
						vform &\rm\it fin\] \\
						val \<\rm\it \@1, \@2\>  \]\\
				arg-st  \<	\@1NP$_{[\mathit{nom},\overrightarrow{\mbox{\tiny {\emph{student}}}}]}$,
							\@2NP$_{[\mathit{acc},\overrightarrow{\mbox{\tiny {\emph{book}}}}]}$
                			\>\\
				sem  \< \[\asort{read-fr} 
						reader  &\rm\it \@1 \\
						theme	 &\rm\it \@2 \]\>
                \]
		\end{avm}
	\end{center}
\caption{Compact representation of \emph{student-read} event}
\label{compact-student-read}	
\end{figure}



\subsection{The Question of Inheritance}

Inheritance plays a central role in Sign-Based Construction Grammars (SBCG). Every construction, regardless of whether it is lexical or combinatory, is associated with its own set of specific constraints. Combinatory constructions are generic in that they are not lexicalized; they correspond to sign types at various levels. More precisely, a single sign may belong to several different types simultaneously, a phenomenon known as multiple inheritance. In this way, a construction can be understood as the aggregation of a collection of generic constructions.

Combined with the principle of locality inherent to SBCG (which establishes the relationship between a mother and its daughter), inheritance enables an incremental, compositional mechanism in which each construction contributes a specific element of meaning that is combined with others. This mechanism operates rather horizontally: unlike derivational approaches that build a hierarchical structure of constituents and propagate information through that hierarchy, SBCG activates multiple sources of information at once, allowing them to interact through multiple inheritance.

As we have seen, in some instances fully saturated constructions, allowing for a global and direct access to a complex semantic element, are activated. In these cases, because the entire construction is directly activated rather than incrementally built, the propagation of information via inheritance is not fully engaged. For example, an idiomatic construction may grant immediate access to a global meaning in a non-compositional manner while still exhibiting variability in its surface form. In such instances, one observes semantic non-compositionality alongside structural compositionality, since the syntactic form remains composable. Here, the inheritance mechanism applies only to the form.

Interestingly, neurolinguistic experiments indicate that when such a construction is activated (as in the case of an idiom), there is no detectable trace of the semantic content associated with an individual word in the brain signal \citep{Rommers13}. In other words, once the surface form is verified, its semantic processing is not carried out compositionally.

\section{Operationalization: The Processing Mechanisms}
Operationalizing Construction Grammars is a complex challenge. One major reason is that the formalization of the theory remained underdeveloped for a long time. Yet a precise formalization is a necessary prerequisite for any computational implementation. Three main approaches have addressed this issue: Sign-Based Construction Grammar (SBCG) \citep{Sag12}, Embodied Construction Grammar (ECG) \citep{BergenChang2005}, and Fluid Construction Grammar (FCG) \citep{Steels06}.

ECG proposes a cognitively inspired approach in which constructions are represented as schemas, with the semantic component grounded in conceptual graphs. In the case of FCG, early efforts were based on inference-rule systems. More recently, a robust development environment has been introduced to support the implementation of grammars within this framework \citep{vanTrijp22}. This approach, where constructions are structured with both a production and a comprehension side, has been particularly useful for studying language learning and evolution.

The contribution of SBCG is especially significant. It builds on the highly detailed formal apparatus of Head-Driven Phrase Structure Grammar (HPSG), adopting its core principles and integrating them into Construction Grammar. Several HPSG implementations have been proposed, some of which have been at least partially adapted to SBCG. From an operational standpoint, this approach is constraint-based, relying fundamentally on two mechanisms: unification and constraint satisfaction.

Our own proposal is fully embedded in this framework. We systematize the use of constraints through what we term properties, enabling us to offer both a precise formal system and an operational implementation. This supports the key mechanisms at the core of our approach: activation and unification.

\subsection{The Activation Function}

Language processing is considered to be a typical predictive mechanism \citep{pickering2013integrated, Pickering21}. Among the different aspects of prediction, the notion of activation plays a central role by making it possible to predict dynamically, in real time, different elements of the meaning, starting from the context. This notion has been explored in many works from different perspectives. In particular, the ACT-R cognitive architecture \citep{Anderson04} has been adapted to language processing  model by proposing a specific activation function \citep{Lewis05}. In this memory-based architecture, several partial structures (or chunks) are stored in working memory. Depending on their level of activation, they can be retrieved and integrated into the structure. The activation level depends on the chunk history (the number of times it has been used and the delay from its last access) and the strength of the relations (the linguistic cues) that link the chunk with the current item. We propose to apply this notion of activation to the prediction of entire pieces of meaning and more precisely to the identification of the nature of the cues and of the ways in which they activate constructions, events and frames forming the core of \emph{Distributional Construction Grammar}.

Activation of linguistic objects can occur at many different levels. Psycholinguistic works have integrated this aspect by proposing a specific language processing architecture \citep{Lewis05,Jager2015,Engelmann16}. In this approach, the different linguistic objects are described in terms of features that can play the role of cues. More precisely, words are encoded in memory as bundles of features, and retrieval cues are used to identify the correct chunk from memory. If retrieval cues match with the features of a chunk in memory, the chunk gets a boost in activation. What is interesting with this approach is that the activation can be evaluated thanks to a predictive function defined as follows:

\begin{equation}
	A_{i} = B_{i} + \sum_{j}{W_{j}S_{ji}}
\end{equation}

According to \citet{Lewis05}, the total activation of a chunk is the sum of its \emph{basic activation B} (frequency and history of the chunk access) and the \emph{Spreading activation S} received from all matching cues $j$, that is a function of the associative strength $S_{ji}$ between cue $j$  and item $i$ (i.e., the probability of the item $i$ being needed given cue $j$) and the cue's weight $W_{j}$:

\begin{equation}
			S_{i} = \sum_{c}{W_{c}S_{ci}}
\end{equation}

The \emph{associative strength} ($S_{ci}$) is the strength of the association $S_{ci}$ between cue $c$ and item $i$, reflecting the probability of the item to be realized given the cue $c$: 

\begin{equation}
	S_{ci} = \mathit{MAS} - ln(\mathit{fan}_{ci})
\end{equation}
In this formula, $\mathit{MAS}$ is the \textit{maximal associative strength} and $\mathit{fan}_{ci}$ is the \textit{fan effect}: the more competitor items match cue $c$, the weaker the association of $c$ with item $i$.

As stated above, the comprehension of a sentence can be modeled as an \textit{incremental process} whose goal is to construct a coherent representation of the events the speaker intends to communicate. In this process, a central role is given to \textit{prediction}. \cite{Pickering21} assume that the central role of both production and comprehension is perceptual prediction (i.e., predicting the perceptual outcomes of an action).  This evidence is compatible with different approaches such as probabilistic models of language comprehension \citep{hale2006,levy2008}, models of complexity that incorporate prediction \citep{Gibson98}, simple recurrent networks \citep{elman1991}, and so on.

Studies on processing difficulty formalize prediction as a \textit{surprisal} measure for the words in a sentence. For instance,  the model proposed by \cite{Hale01} predicts the difficulty of integrating a new word into a syntactic structure as the negative logarithm of the probability of a word given its previous linguistic context. However, syntactic effects only account for a small part of the cognitive load in sentence processing, while semantic information from sources as \emph{Generalized Event Knowledge} \citep{McRae09}, \emph{Situation Model} \citep{Zwaan98}, \emph{Common Ground} \citep{Stalnaker02} 
plays a major role by restricting the universe of interpretation.




Prediction can be at different levels: from simple priming 
to activation, which creates an expectation (e.g., filling a complement, an argument value, etc.). Prediction is done by evaluating or satisfying a set of constraints, such as exclusion, co-occurrence restriction, requirement, lexical selection, and so on. This kind of expectation can be quantified in terms of an activation score. According to \cite{Lewis05}, the total activation of a chunk \textit{i} is a function of the sum of its \textit{base activation} (frequency and history of \textit{i}) and the \textit{spreading activation}, described as  the product for each cue \textit{c} between \textit{c}'s weight and the associative strength between \textit{i} and \textit{c}. The total activation of a chunk determines both retrieval latency and probability of retrieval.

Since constructions are form-meaning pairings, the activation function has to be modified in order to include different cues (morphological, syntactical, distributional, etc.) that give access to the construction.
The problem is thus determining i.) the complete set of cues that activates a \emph{Cx}, which can be features, conjunction of features or word forms that activate all \emph{Cx}, and ii.) how to weight cues, distinguishing between mandatory vs. optional cues (or hard vs. soft ones). Among the cues that could activate a \emph{Cx}, lexical  ones are particularly crucial.  Lexical cues are distributional semantics vectors: The spreading estimation of activation should also include the evaluation of the distance between the vector of the word in the construction  and that of the prototype encoded into the representation of the event. 

Accordingly, the spreading activation $S_{ji}$ introduced above should be re-written in the following direction:
\begin{equation}
	A_{i} = B_{i} + \sum_{j}{W_{j}F_{j}S_{ji}}
\end{equation}

\noindent{}In this formula,  $F_{j}$ is the evaluation of the distance between the word vector and that of the prototype. It is thus necessary to define the distance measure  and the features to take into account. Moreover, cues not only contribute to activation evaluation, they are also a way to control the instantiation of the constructions by specifying an adequate unification mechanism, involving distributional information.

We propose an activation score expressing the salience of the object as well as the strength with which it is activated by cues. In a first approach, \cite{chersoni2016towards,Chersoni17,chersoni2019structured,chersoni2021not} proposed a function of the semantic coherence of the unified event and of the salience of the cues:\footnote{In \citet{chersoni2016towards,chersoni2019structured}, the authors experimented on the Bicknell dataset \citep{bicknell2010effects}, which is composed of SVO sentence pairs where the final direct object is more or less plausible depending on the subject (e.g. \textit{The journalist was fixing the spelling} vs. \textit{The mechanic was fixing the engine}), and the task of a model is to identify the most typical sentence in each pair. The models in \citet{chersoni2019structured,chersoni2021not} tackle the task by making use of an activation function to compute how strongly verb and object are activated by the subject, via corpus-extracted probabilities. The framework could ideally be updated using Surprisal computed by modern LLMs, because typical argument combinations would probably lead to lower Surprisal values.}

 \begin{itemize}
 	\item \emph{Semantic coherence} $\theta$, that weights the internal semantic coherence of a \textit{Cx} as the typicality (thematic fit) of its components. Generally, the higher is the mutual typicality between the components, the higher is its internal semantic coherence. In order to compute the semantic coherence of a \textit{Cx}, we can rely on the distributional information encoded in the {\sc ds-vector} feature. That is, the thematic fit of \textit{student} as a subject of \textit{read} is given by the cosine between the vector of \textit{student} and the prototype vector built out of the \textit{n} most salient subjects of the verb \textit{read} \citep{Baroni10,lenci2011composing}.
 
 	\item \emph{Salience} $\sigma$: function of the cue weights, the strength with which it is activated (cued) by the composed linguistic expressions. This entails that events that are cued by more linguistic constructions in a sentence should incrementally increase their salience
\end{itemize}

The scores are calculated while building the situation model (i.e., when instantiating the object with realized arguments). The final interpretation of a linguistic input \textit{l} therefore amounts to identifying the object described by recognizing the cues brought by various lexical and syntactic constructions in \textit{l}. Interpreting a linguistic input relies on different processes, between memory retrieval and composition: when the context is very familiar (i.e.,  highly predictable frames and events), the meaning can be already stored in our long term memory, and then it is identified via the linguistic cues. However, in other cases it can be a brand new situation which we have never encountered before, which we must then build by means of compositional processes.

An alternative is represented by current probabilistic models of language comprehension, which are generally based on the Surprisal theory framework \citep{Hale01}, 
claiming that the effort a reader must spend to comprehend a word is inversely related to its contextual predictability.
Given the surprisal values estimated by a Large Language Model (e.g., GPT-2), a large body of literature analyzed the relationship between surprisal and human reading times and found strong supporting evidence for the theory \citep{smith2013effect,wilcox2023testing,shain2024large}. A shortcoming of surprisal is that words and constructions can be contextually predictable for different reasons, and probabilities (on which the metric is based) conflate them together. However, it has also recently been proposed that similarity might also play a role in processing difficulty \citep{salicchi2023study,meister2024towards}. In particular, \cite{meister2024towards} points out that several words may be activated by a linguistic context because they share some similarities at the syntactic, semantic or orthographic level, and propose a framework to incorporate different types of similarity metrics in the computation of the surprisals for words. We found such a proposal in line with the spirit of our framework. One could re-weight surprisal scores with word embedding cosines to estimate activations and facilitation effects due to semantic constraints (e.g., coherence constraints in verb-argument constructions), or to use boolean values to account for the satisfaction or violation of grammatical constraints specified in the \textsc{properties} feature. In other words, we see the Surprisal framework as compatible with our proposal, as long as surprisals can be combined with the evaluation output of more specific linguistic constraints.

An additional concern could be that the segmentation of most Large Language Models does not align with the meaningful morphological units of the language, e.g. typical tokenization schemes such as BPE generally split a word into subwords, and this would make the mathematical operationalization at the morphological level problematic \footnote{But see also the recent findings of \citet{ciaccio2025beyond}, showing the emergence of character-level and, to some extent, morpheme-level knowledge in Large Language Models as training data and parameter size increase (the so-called \textit{spelling miracle}, \citet{liu2022character}).}. However, we argue that subword tokenization is not a necessary feature of language models: for example, the work of \citet{nair2023words} shows how a simple language model trained on the output of an English morphological transducer achieves a better fit with human reading times, and particularly on the words that would be normally split by a subword-based tokenizer. We take this study as preliminary evidence that, when pretraining is based on a linguistically-motivated tokenization scheme, surprisals can be used as a realistic estimate of human language processing difficulty.

\subsection{Types of Cues, Types of Activation}

Cues can be of different nature, depending on the type of the object to be activated (frame, event, construction), each being potentially associated to a specific activation mechanism.

\subsubsection{Lexical Cues: Similarity}

In most of the examples (including those presented above), cues are considered as being lexical forms. The idea is that some words, when occurring in an input, activate structures. All three types of \textit{DCxG} structures can be activated by lexical items. 

Frames can be \textit{evoked} by lexical cues,  which are form/meaning pairings associated to words. This notion is similar to activation, the general idea is that a lexical cue can play the role of a cue, activating an entire frame. An illustration is given in \cite{ruppenhofer2016framenet}: ``\emph{for example, the {\sc Apply\_heat} frame describes a common situation involving a {\sc Cook}, some {\sc Food}, and a {\sc Heating\_Instrument}, and is evoked by words such as \emph{bake, blanch, boil, broil, brown, simmer, steam}, etc.}''. Each frame comes with a set of lexical units. The intuition is that the realization in the context of one or several lexical cues pertaining to the frame description activates it. We propose to add a new feature in the frame description stipulating explicitly what are lexical units that can be considered as a lexical cue (see Figure \ref{lex-cues}). 

\begin{figure}
		\begin{center}
            \avmsortfont{\it}
            \avmvalfont{\it}
            \begin{avm}
            \[\asort{commercial-transaction-frame} 
                sem & \< \[
                			buyer   & $\overrightarrow{\mathtt{buyer}}$ \\
                			seller	& $\overrightarrow{\mathtt{seller}}$ \\
							goods	& \[ds-vec $\overrightarrow{\mathtt{goods}}$ \\ price \rm\em p\] \\
							money	& \[ds-vec $\overrightarrow{\mathtt{money}}$ \\ value \rm\em p\] \\
							place	& $\overrightarrow{\mathtt{shop}}$ \\
            				\]\>  \\
				 cues & \[ lex-cues \rm\it \{\rm\it{$\overrightarrow{\mathtt{shop}}$, $\overrightarrow{\mathtt{shopkeeper}}$, $\overrightarrow{\mathtt{good}}$, $\overrightarrow{\mathtt{pay}}$}, ... \}\]
			\]	
            \end{avm}
		\end{center}
\caption{Commercial transaction frame with cues}
\label{lex-cues}
\end{figure}

In \emph{DCxG}, the first step of the activation evaluation consists in verifying the matching between the lexical items in the input and the lexical units described in the frame. As stated above, this matching is done thanks to a double process: \textit{similarity} and \textit{unification}. Considering words as cues means that the realization of some specific lexical forms in the context entails the activation of an object. In our case, in line with  \emph{Frame Semantics}, the realization of some words can then \textit{activate} a frame. Concretely, when a lexical unit in the input is similar to one in the frame lexical cues, the corresponding frame is activated. Activating a frame does not mean that it will be considered as instantiated, but simply that it becomes accessible, and gets added to the list of preferred frames among others. As indicated in the activation formula, the level of activation increases with the number of cues that are instantiated. In the case of frames, the more lexical units in the frame (i.e., its cues) are realized, the higher its activation (representing its predictability) will be.

\subsubsection{Syntactic Cues: Unification}

Similarity thus plays  a major role in the evaluation of the activation level. But unification still constitutes a way to identify the matching between input objects (events, frames, or constructions) and their abstract description in the grammar. 

In this case, it is possible to mention only some specific feature values that could play the role of cues. This is the reason why the {\sc lex-cues}, as illustrated in Figure \ref{lex-cues}, are not distributional vectors but complete lexical objects, that bear not only their distributional information but also potentially other features. For example, in the case of the \emph{subject-predicate construction} (presented above in Figure \ref{abstract-cx}), the morphological feature \emph{verb form} indicates that this construction applies to all finite verbs that are not auxiliary and not inverted (whatever the value of the other features). In this example, as soon as this verb form value is instantiated, the construction is activated. This indicates that the morphological value plays the role of a cue for the \emph{subject-predicate construction}, as specified in the complete version of the construction in Figure \ref{subj-pred-final}.
Any feature can thus be a cue, even though not all features are cues, and therefore it is necessary to specify this information in the object. Besides lexical forms and morpho-syntactic features, it is then possible to specify other types of features as cues.

\begin{figure}
\begin{center}
    \begin{avm}
 			\[syn \[cat \@1\[vf \rm\it fin \\
        					inv - \\
							aux -\] \\
					mark \rm\it unmk \\
					val \<\@3\> \]  \\
			cues \[syn-cues \@1\]\]
	\end{avm}
\end{center}
\label{subj-pred-final}
\caption{Final version of the \emph{subject-predicate} construction}
\end{figure}

As stated by the \emph{Property Grammars} \citep{Blache16b}, a construction is recognized thanks to its specific form, which relies on a  set of   \textit{properties} (from any domain: morphology, prosody, syntax, etc.). Some properties can be represented as relations between its constituents, as \textit{linearity} (the linear order existing between two words), \textit{co-occurrence} (mandatory co-occurrence between two words), \textit{dependency} (syntactic-semantic dependency between two words), \textit{adjacency} (constituents that have to be in an adjacent position), and so on.  For example, as described above, the ditransitive construction is characterized by some specific properties constraining the realization of the different arguments in a specific order, encoded in the {\sc linearity} and {\sc adjacency} features. In this construction, the order of realization of the argument is the main cue triggering its activation, as represented in Figure \ref{ditrans-cues}.

\begin{figure}
	\begin{center}
        \begin{avm}
        	\[form
        		\[syn \[cat & \@1 V \] \\
			  	  properties \[\@{10}lin \{\@2 NP$_{[subj]}$ $\prec$ \@1 $\prec$ \@3 NP$_{[obl]}$ $\prec$ \@4 NP$_{[obj]}$\} \\
			  		 \@{11}adj \{\@1 $\oplus$ \@3; \@3 $\oplus$ \@4\} \]					
				\] \\
			  arg-st \< \@2, \@3, \@4 \> \\
			  sem \[ frames \<\[ \asort{transfer-fr} 
									agent &\rm\it \@2 \\	
									recipient &\rm\it \@3 \\	
									theme &\rm\it \@4 \]\>\]\\
			  cues \[syn-cues\{\@{10}, \@{11}\}\]
			\]
        \end{avm}						
	\end{center}
\label{ditrans-cues}
\caption{Cues of the ditransitive construction}	
\end{figure}

Note that all cues can be associated to weights, implementing their relative importance. We propose as a first approximation to distinguish between two weighting values: \emph{hard} and \emph{soft}. For example, in the ditransitive construction, the linearity and adjacency constraints between the oblique and object complements are mandatory, whereas the linearity between the subject and the verb is less important. This information can be added to each cue. 

The same type of approach can also be used with more lexicalized (more or less frozen) constructions such as idioms, by mixing both lexical and syntactic cues. For example, the idiom ``\emph{to put all eggs in one basket}'' is mainly characterized by its lexical units and their order (only few variation is possible in most of idiomatic constructions). This information is encoded by the corresponding lexical and syntactic features, that are taken as  cues. Moreover, this representation implicitly encode  an important aspect for idiom recognition: an idiom is usually identified rapidly, after the 2 or 3 first words, called the \textit{recognition point}. Figure \ref{idiom-cues} implements these different aspects, specifying the form and the cues. In this example, the recognition point is reached after the third word: ``\emph{put all eggs} ...'' is recognized as an idiom by native speakers that can complete  the rest of the idiom automatically. 

\begin{figure}
	\begin{center}
        \begin{avm}
        	\emph{idiom} $\Rightarrow$
        	\[form \[surface\_form \<\@1\ \it put; \@2\ \it all; \@3\ \it eggs; \it in; \it basket \> \]\\
			  properties \[ \@{10}lin \{\@1 $\prec$ \@2 $\prec$ \@3\} \\
			  		 		\@{11}adj \{\@1 $\oplus$ \@2; \@2 $\oplus$ \@3\} \]\\
        	  sem \[frames\<\[\asort{take-a-risk-fr}
	  							...\]\>\] \\
			  cues \[lex-cues \{\@1, \@2, \@3 \}\\
					 syn-cues \<\@{10}, \@{11}\>\]					
			\]
        \end{avm}						
	\end{center}
\label{idiom-cues}
\caption{Cues for idiomatic construction}	
\end{figure}

Concretely, when scanning an input sentence containing this idiom, the recognition of the first three words (in other words reaching the recognition point) makes it possible to verify the different constraints stipulated in the cues: all these three words correspond to the lexical cues, the linearity and the adjacency constraints between these lexical units are also satisfied. In such case, all cues being satisfied, the corresponding construction is activated, making it possible to access directly to the idiom meaning. 

 One significant criticism of strict unification-based linguistic models is their dependence on exact feature matching. When two syntactic structures bear incompatible features, traditional unification predicts a parse failure. However, speakers may not always construct detailed syntactic representations on the fly. Instead, they frequently utilize large, unalized chunks of language-surface strings or formulaic patterns that coexist alongside fully parsed structures in memory \citep{Ferreira07,dkabrowska2014recycling}. Such representations can interfere especially when superficially identical but structurally distinct constructions overlap--a phenomenon known as \textit{constructional contamination} \citep{pijpops2018constructional}. Constructional contamination arises when two unrelated constructions yield similar surface forms, leading to probabilistic shifts in the preferred variant of one construction under influence from the other.
By integrating unification with activation-based weighting, our proposed framework would be able to simulate how frequently encountered chunks or form-similar constructions exert gradient influence, allowing structurally divergent yet superficially alike constructions to occasionally interfere in comprehension and production.

\section{Discussion and Conclusion}

In this paper, we have introduced the formal basis of a new architecture that explains how different mechanisms are at play in the comprehension process. 

Our proposal introduces two significant novelties: i) it offers a complete representation of the semantic information integrated into the \textit{Construction Grammar} framework, and ii) it incorporates distributional vectors within the construction representation, thereby accommodating the more usage-based aspects of meaning. On the one hand,  a final linguistic representation is the result of three informational structures (construction, frames, and events) that implement different levels of semantics, from lexical meanings to background knowledge. This organization accommodates explaining why interpretation can stay shallow (only based on generic information) or, on the contrary, detailed (the generic objects being specialized by the specific local and contextual information). On the other, the meaning representations are not hand-coded, but they derive from the distributional statistics observed in text corpora, coherent with the idea that language emerges from language use. The first contribution of the paper is, therefore, theoretical: we offer a new representation of the constructional and semantic information by integrating these different aspects into the formalism provided by \textit{Sign-Based CxG}. We name this new framework \textit{Distributional Construction Grammar} (DisCxG).  

The description offered by DisCxG complements existing representations and computational CxG models (such as \citealp{Dunn2024}). 
Indeed, a growing body of work focuses on learning constructions automatically from corpora. These models leverage unsupervised or semi-supervised learning techniques to extract recurring form-meaning pairings from large-scale language data. The learned constructions are then organized into structured networks, often modeled as graphs, which capture grammatical generalizations across levels (e.g., syntax, semantics, discourse). 
The aim of this line of research is both theoretical (to demonstrate that cxgs are learnable from usage data) and applicative (automatically filling a construction would make the work scalable to a larger number of expressions, instead of having small human-annotated resources).
While this research direction represents a crucial advancements, especially in terms of descriptive coverage and scalability, it still focus on static aspects of grammar, leaving unexplored how constructions function dynamically during actual language processing--that is, how constructions are activated, composed, and interpreted in real-time comprehension and meaning construction.

The framework offers a complementary perspective: DisCxG formalism is designed to investigate the interaction between symbolic constructional knowledge and distributional semantics during sentence processing. In other words, besides the formal contribution of DisCxG, this paper aims to represent the first attempt towards an implementation of how the compositional and non-compositional routes can be integrated into a unique framework. Specifically, we presented different mechanisms for implementing comprehension processing: i) we specify the role of the different features describing semantic information, ii) we explain the interaction between the three informational structures, and iii) we describe the basic mechanisms to access meaning, respectively activation, similarity, and unification. These processing mechanisms open the door to a new view of language processing, taking advantage of different operational paradigms (numerical and symbolic) and integrating both compositional and non-compositional approaches.


Future works around the \textit{Distributional Construction Grammar} architecture will address different directions. First, corpus-based studies will offer the possibility to precise the activation function and the integration of similarity into the unification mechanism. Second, several behavioral experiments have validated the double-route hypothesis by contrasting when one mechanism is at play instead of another (cf. \citealt{carrol2021psycholinguistic} for an overview on idioms and figurative language). The data from these experiments could be used as a testing ground for computational modeling, to see if the predictions of our architecture are reflective of the human behavior observed in the experiments (similarly to \citealp{MichalonBaggio2019}). 
Finally, another research track will be to investigate cognitive aspects, for example, by evaluating the compatibility of this framework with neuro-cognitive architectures such as \textit{Memory, Unification, and Control} \citep{Hagoort13,Blache24}.

A final consideration regards how this work can fit into the recent advancements in linguistics and AI. The rise of large language models (LLMs) invites important questions about how constructional knowledge might be represented and processed in this new paradigm \citep{goldberg2024usage, madabushi2025construction}.

A key challenge lies in interpreting how constructional knowledge is encoded within these high-dimensional representations, and whether models can reliably distinguish subtle differences in form-meaning pairings. Recent works show that LLMs implicitly capture constructional patterns, but struggle to grasp the semantics of more schematic constructions \citep{weissweiler-etal-2022-better, madabushi2020cxgbert}. This is exactly the reason why we argue for the importance of a more structured representation of semantics. Moreover, while current investigations examine whether LLMs can recognize or replicate known constructional distinctions, less attention has been paid to how such knowledge is deployed during interpretation and production. In contrast, our goal is to propose a framework that is not only descriptively adequate but also cognitively plausible, thereby narrowing the gap between CxG and psycholinguistic theories of language processing.

Our framework represents an initial step toward more cognitively grounded formalizations of Construction Grammar. We hope that this work encourages deeper collaboration between computational linguists and constructionist theorists, facilitating the development of a scalable, efficient CxG formalism that integrates usage-based information and structured representations.



\bibliographystyle{apalike}
\bibliography{Ref-DCxG.bib}

\end{document}